\documentclass[letterpaper]{article} % DO NOT CHANGE THIS
\usepackage{aaai25}  % DO NOT CHANGE THIS
\usepackage{times}  % DO NOT CHANGE THIS
\usepackage{helvet}  % DO NOT CHANGE THIS
\usepackage{courier}  % DO NOT CHANGE THIS
\usepackage[hyphens]{url}  % DO NOT CHANGE THIS
\usepackage{graphicx} % DO NOT CHANGE THIS
\usepackage{natbib}  % DO NOT CHANGE THIS AND DO NOT ADD ANY OPTIONS TO IT
\usepackage{caption} % DO NOT CHANGE THIS AND DO NOT ADD ANY OPTIONS TO IT
\usepackage{algorithm}
\usepackage{algorithmic}

\usepackage{newfloat}
\usepackage{listings}
\DeclareCaptionStyle{ruled}{labelfont=normalfont,labelsep=colon,strut=off} % DO NOT CHANGE THIS
\floatstyle{ruled}
\newfloat{listing}{tb}{lst}{}
\floatname{listing}{Listing}
\title{\method:  Aligning Large Multimodal Models for Videos by Iterative Self-Retrospective DPO}
\author{
    Daechul Ahn\textsuperscript{\rm 1}\equalcontrib,
    Yura Choi\textsuperscript{\rm 1,2}\equalcontrib,
    San Kim\textsuperscript{\rm 1},
    Youngjae Yu\textsuperscript{\rm 2},
    Dongyeop Kang\textsuperscript{\rm 3},
    Jonghyun Choi\textsuperscript{\rm 1}\thanks{JC is with ECE, ASRI and IPAI in SNU and a corresponding author.}
}
\affiliations{
    \textsuperscript{\rm 1}Seoul National University\hspace{0.3em}
    \textsuperscript{\rm 2}Yonsei University\hspace{0.3em}
    \textsuperscript{\rm 3}University of Minnesota\\
    \vspace{0.2em}
    \{daechulahn,00sankim,jonghyunchoi\}@snu.ac.kr\hspace{0.2em}
    \{yoorachoi,yjy\}@yonsei.ac.kr\hspace{0.2em}
    dongyeop@umn.edu
}

\newcommand{\specialcell}[2][c]{%
  \begin{tabular}[#1]{@{}c@{}}#2\end{tabular}}

\def\partial{\textcolor{blue}{\textbf{partial}}}

\usepackage{multicol}
\usepackage{multirow}
\usepackage{booktabs}
\usepackage{pifont}
\usepackage{xcolor}

\usepackage{amsmath}
\usepackage{xspace}
\usepackage{colortbl}
\usepackage{amsfonts}
\usepackage{amsmath}
\usepackage{enumitem}

\makeatletter
\DeclareRobustCommand\onedot{\futurelet\@let@token\@onedot}
\def\@onedot{\ifx\@let@token.\else.\null\fi\xspace}

\def\eg{\emph{e.g}\onedot} 
\def\ie{\emph{i.e}\onedot} 
 
 \def\vs{\emph{vs}\onedot}

\makeatother

\newcommand{\method}{\mbox{\textsc{ISR-DPO}}\xspace}

\begin{document}

\maketitle

\begin{abstract}
Iterative self-improvement, a concept extending beyond personal growth, has found powerful applications in machine learning, particularly in transforming weak models into strong ones. 
While recent advances in natural language processing have shown its efficacy through iterative preference optimization, applying this approach to Video Large Multimodal Models (VLMMs) remains challenging due to modality misalignment.
VLMMs struggle with this misalignment during iterative preference modeling, as the self-judge model often prioritizes linguistic knowledge over visual information. 
Additionally, iterative preference optimization can lead to visually hallucinated verbose responses due to length bias within the self-rewarding cycle.
To address these issues, we propose Iterative Self-Retrospective Direct Preference Optimization (\method), a method that uses self-retrospection to enhance preference modeling.
This approach enhances the self-judge's focus on informative video regions, resulting in more visually grounded preferences. 
In extensive empirical evaluations across diverse video question answering benchmarks, the \method significantly outperforms the state of the art.
We are committed to open-sourcing our code, models, and datasets to encourage further investigation.
\url{https://github.com/snumprlab/ISR-DPO}
\end{abstract}

\begin{figure}[t]
    \centering
    \includegraphics[width=0.9\linewidth]{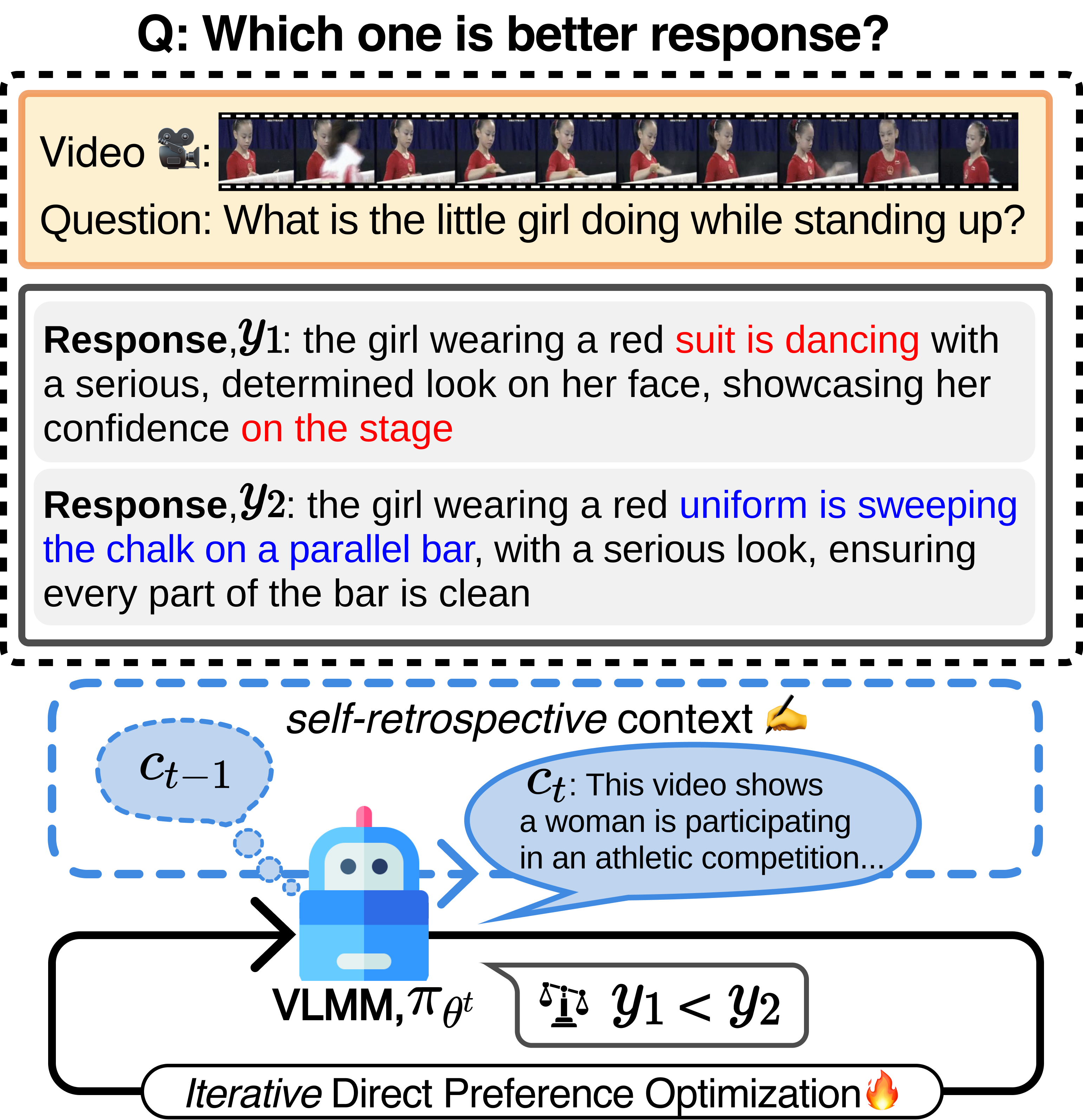}
    \caption{\textbf{Illustration of the proposed \method.} 
    During iterative direct preference optimization (DPO) in VLMM, we select preferences from responses based on not only video content but also visual context $c_t$, \ie, detailed video description, to ensure preferences are grounded in video information. Specifically, we enhance the context in the self-retrospective manner by leveraging context $c_{t-1}$ generated in previous iteration, a process we call \emph{self-retrospective} preference modeling.
    \textcolor{red}{Red} indicates irrelevant responses, while \textcolor{blue}{blue} indicates accurate, visually-grounded responses.
    }
    \label{fig:teaser}
    \vspace{-0.5em}
\end{figure}

%%%%%%%%%%%%%%%%%%%%%%%%%%%%%%%%%%%%%%%%%%%%%%%%%%%%%%%%%%%%%%%%%%%%%%%%%%%%%%%%%%%
\section{Introduction}

\begin{quote}
\textit{Progress is not achieved by luck or accident, but by working on yourself daily.}
\begin{flushright}
--- Epictetus
\end{flushright}
\end{quote}

The human capacity for growth through consistent effort and repetition is a fundamental principle of personal development~\cite{dweck2006mindset}. 
This concept of iterative self-improvement extends beyond personal growth, finding powerful applications in machine learning to transform weak models into strong ones, without relying on additional human-annotated training data~\cite{schapire1990strength,yuan2024selfrewarding,burns2023weaktostronggeneralizationelicitingstrong}.
Notably, recent advances in natural language processing (NLP) have demonstrated the efficacy of \emph{iterative} preference optimization in aligning Large Language Models (LLMs) with human intentions~\cite{yuan2024selfrewarding,pang2024iterative,chen2024selfplay}. 
This approach involves constructing increasingly informative preferences through iterative preference modeling, \ie, LLM-as-a-judge, leading to progressively better-aligned models.

\begin{figure}[!t]
    \centering
    \includegraphics[width=\linewidth]{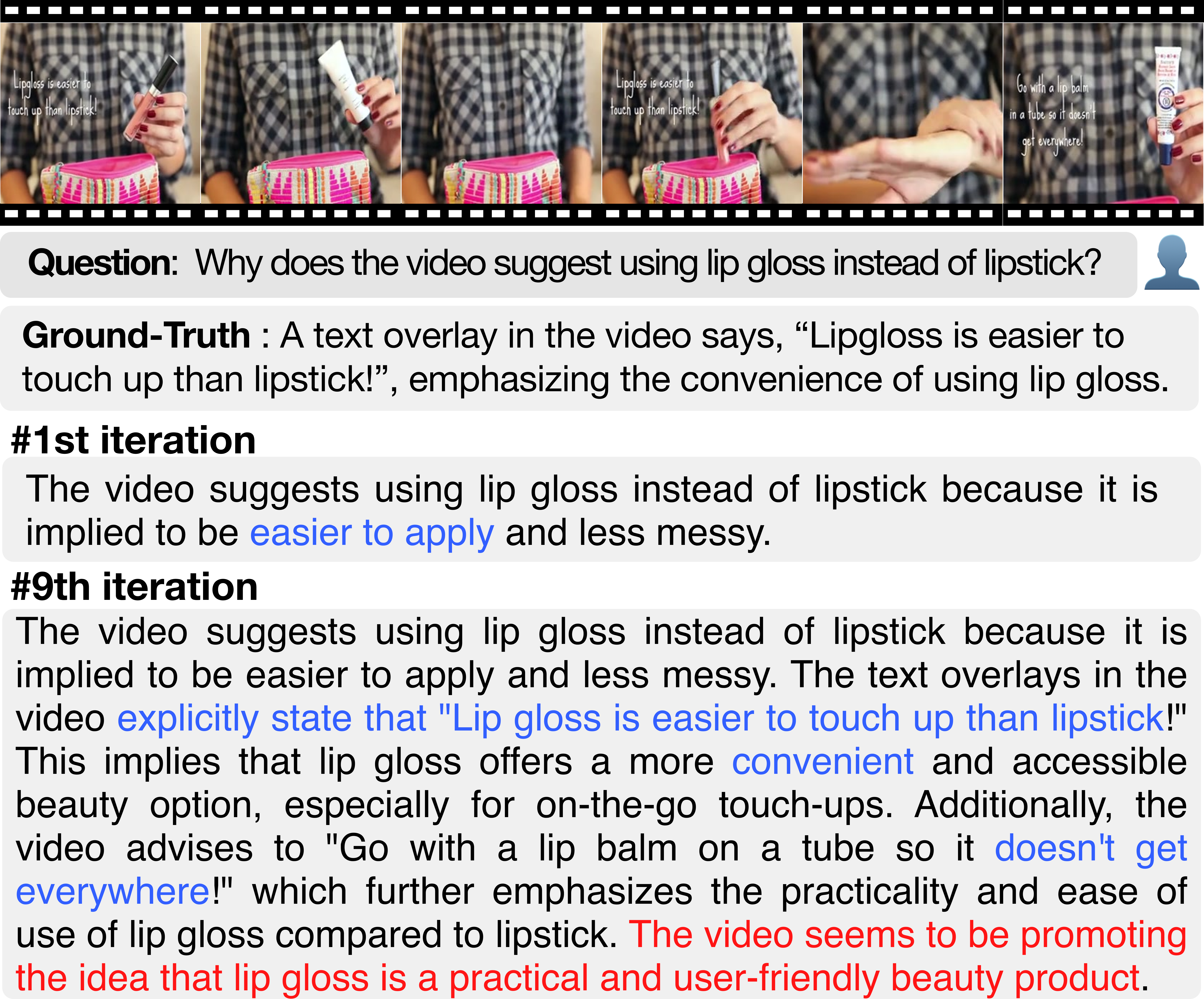}
    \caption{\textbf{Example of verbosity hallucination within iterative preference modeling cycle for VLMM.} 
    At the 1st iteration, the response is concise and visually grounded (in \textcolor{blue}{blue}). By the 9$th$ iteration, the response elaborates further, referencing explicit text overlays in the video. However, it starts to include irrelevant details and assumptions as well, leading to \emph{verbosity hallucination} highlighted in \textcolor{red}{red}.
    }
    \label{fig:teaser2_example}
\end{figure}

However, this iterative self-improvement principle for LLMs poses specific challenges when applied to large multimodal models, particularly Video Large Multimodal Models (VLMMs).
VLMMs suffer from modality misalignment during iterative preference modeling, where the self-judge model tends to rely more on their pre-existing linguistic knowledge rather than the given visual information~\cite{vlm_rlaif,zhou2024calibratedselfrewardingvisionlanguage}.
This leads to preference data that are linguistically plausible but less grounded in visual content.
Moreover, iterative training exacerbates the visually \emph{ungrounded} verbose response in VLMMs due to the length bias within the iterative preference modeling cycle, which favors linguistically \emph{longer} response during preference selection~\cite{rlhflengthbiases,park2024disentangling}. 
As illustrated in Fig.~\ref{fig:teaser2_example}, while somewhat longer responses might enhance the quality of the predicted response, excessively long responses can introduce content irrelevant to the actual video or question, \ie, \emph{verbosity hallucination}, without necessarily improving quality.

\begin{figure*}[t]
    \centering
    \includegraphics[width=0.95\linewidth]{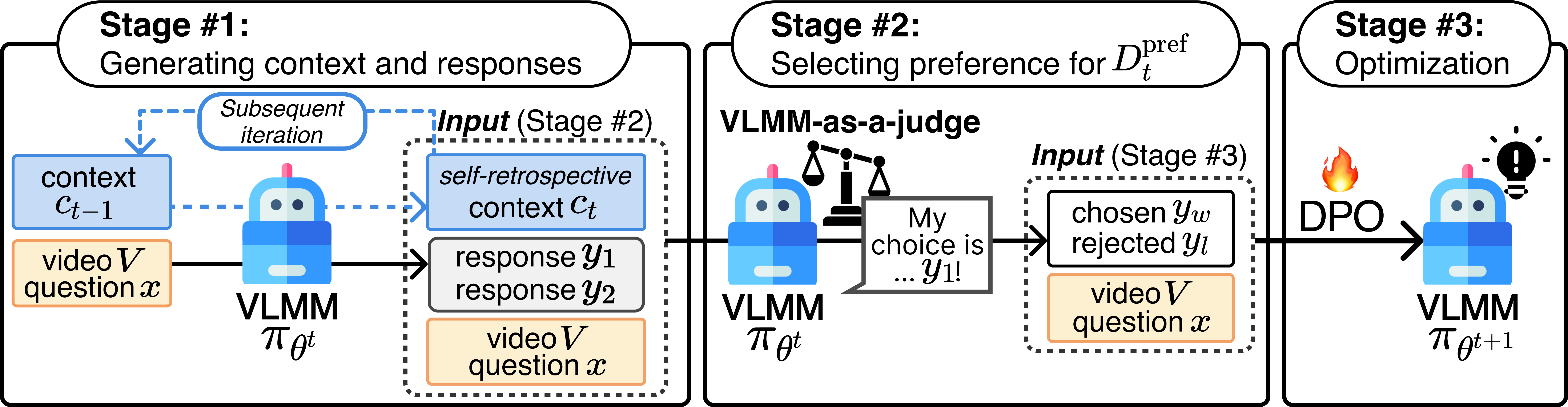}
    \caption{\textbf{Overview of self-retrospective Direct Preference Optimization (DPO).} 
    Each iteration of \method involves three stages: 1) After training iteration $t$, the latest updated VLMM ($\pi_{\theta^{t}}$) generates two different responses $y_1$ and $y_2$ for the given video $V$ and instruction $x$. In addition, a visual description, \ie, visual context, is generated through self-retrospection, providing the necessary input for the next stage, as indicated by the black dotted line. 2) Using the information generated in the previous stage, the model ($\pi_{\theta^{t}}$) compares its responses($y_1$ and $y_2$) and classifies the preferred response $y_w$ and the rejected response $y_l$.
    3) Then, the VLMM ($\pi_{\theta^{t}}$) is optimized using DPO to update the parameters to $\pi_{\theta^{t+1}}$. 
    }
    \label{fig:overview}
\end{figure*}

To address these challenges, we argue that the self-judge model, \ie, VLMM, should select preferences based on visual content, rather than being merely linguistically plausible at each iteration.
We achieve this visually grounded self-judgment by drawing inspiration from cognitive science on human perception~\cite{Bransford1972,Kintsch1988,Anderson1984}, emphasizing the importance of contextual information in interpreting visual data.
Specifically, we provide the self-judge with additional video descriptions generated through a self-retrospective manner as an additional visual context.
This additional information acts as a focusing mechanism, akin to attention in human cognition~\cite{Bransford1972}, enabling the VLMM to ground its responses more effectively in the video, reducing the likelihood of generating irrelevant or hallucinated one.

To this end, we propose a simple yet effective iterative self-improvement approach for VLMM: \textbf{I}terative \textbf{S}elf-\textbf{R}etrospective \textbf{D}irect \textbf{P}reference \textbf{O}ptimization (\method) as shown in Fig.~\ref{fig:teaser}.
This approach helps the self-judge focus on more informative regions in the video when comparing responses, producing more visually grounded preferences at each iteration.
Our empirical studies demonstrate that our \method exhibits superior performance compared to state-of-the-art VLMMs on various video question answering benchmarks.

We summarize our contributions as follows:
\begin{itemize}
\setlength\itemsep{-0.2em}
    \item We propose a novel modality alignment method for video large multimodal models (VLMMs), utilizing iterative direct preference optimization (DPO) to align video-text modalities effectively.
    \item We enhance AI's feedback by proposing self-retrospective preference modeling, which improves clarity and comprehension in video through the use of iteratively refined visual context for preference selection.
    \item We demonstrate the effectiveness of our proposed \method on various video question answering benchmarks by a noticeable margin.
\end{itemize}

%%%%%%%%%%%%%%%%%%%%%%%%%%%%%%%%%%%%%%%%%%%%%%%%%%%%%%%%%%%%%%%%%%%%%%%%%%%%%%%%%%%
\section{Related Work}
\paragraph{Aligning large multimodal models for videos.}
VLMMs have achieved notable success in various video comprehension tasks, such as video temporal understanding~\cite{liu2023one}, question answering~\cite{videollava}, and instruction-following~\cite{Maaz2023VideoChatGPT}. 
These models integrate publicly available LLMs~\cite{touvron2023llama,touvron2023llama2} with visual encoders~\cite{clip} and additional learnable parameters~\cite{hu2022lora}, undergoing Supervised Fine-Tuning (SFT)~\cite{Maaz2023VideoChatGPT,videollava,li2023llamavid} and, more recently, preference optimization~\cite{rafailov2023dpo,llava-hound,vlm_rlaif}.
Our work builds upon these efforts by exploring the application of iterative preference optimization to VLMMs and addressing the unique challenges related to length bias and visual grounding during preference modeling process.

\paragraph{Iterative preference optimization.}
Training LLMs with preference optimization has proven to be an effective approach to align language models with human intention, improving model performance and reliability.
Build upon this preference optimization, recent efforts have focused on iterative preference optimization techniques, which typically involve iteratively generating feedback data with AI models themselves, \ie, \emph{self-rewarding}.
Many recent work in the NLP domain concurrently propose this, where the aligned model iteratively generates responses and judges its own outputs to build feedback data and learn from this data with DPO~\cite{yuan2024selfrewarding,pang2024iterative,chen2024selfplay}. 
While these iterative optimization techniques have shown their effectiveness in LLMs, their application in the multimodal domain, particularly for video understanding tasks, remains largely unexplored.
Our work proposes an effective iterative preference optimization method for VLMMs.

\paragraph{Verbosity bias in preference optimization.}
Preference fine-tuning methods such as RLHF, RLAIF, and DPO are known to produce responses that are longer than those generated prior to preference optimization, known as length bias. 
This phenomenon stems from a verbosity bias in preference data, where both human and AI judges tend to favor longer responses~\cite{rlhflengthbiases,park2024disentangling,saito2023verbosity}.
Despite minimal differences in length between preferred and rejected responses, the increase in verbosity is statistically significant~\cite{park2024disentangling}.
In VLMMs, this length bias can be particularly problematic.
It may result in verbose responses that are linguistically comprehensible but not well-grounded in the visual content. 
Addressing length bias in the multimodal setting of VLMMs remains an open challenge.

%%%%%%%%%%%%%%%%%%%%%%%%%%%%%%%%%%%%%%%%%%%%%%%%%%%%%%%%%%%%%%%%%%%%%%%%%%%%%%%%%%%
\section{Iterative Self-Retrospective DPO}
To effectively align the multimodalities between video and text, we propose to use an iterative self-improvement approach for VLMM.
Figure~\ref{fig:overview} illustrates the overall training pipeline of our proposed \method for one cycle, which executes three stages: 1) generating self-retrospective context and responses, 2) selecting preferences, and 3) optimization.

During iterative execution, we enhance our model's ability to select preferences by conditioning not only the video content, but also on the visual context generated through self-retrospection.
This additional visual context generates preferences grounded in the video, improving the alignment between visual and textual modalities.

\subsection{Iterative DPO in VLMM}
\label{sec:approach:iter_dpo}

We denote the current VLMM at the $t$-th iteration as $\pi_{\theta^{t}}$. 
This model generates responses and selects preferences by itself, thereby constructing the preference data, $D_{t}^{\text{pref}}$.  
With $D_{t}^{\text{pref}}$, we train the subsequent VLMM, denoted as $\pi_{\theta^{t+1}}$, at the $t+1$-th iteration.

\paragraph{Initial model.}
Given a seed preference data annotated in~\citet{llava-hound}, we conduct preference fine-tuning using DPO, starting from the SFT model provided from previous work~\cite{llava-hound}.
This preference fine-tuned model is referred to as the initial model $\pi_{\theta^{1}}$.

\paragraph{Preference modeling.}
Given the current VLMM $\pi_{\theta^{t}}$, we generate two different responses for the input video $V$ and question $x$ using a high temperature hyper-parameter (\eg, 0.7).
This high temperature flattens the token sampling probability distribution, producing varied responses from the same input in the current VLMM $\pi_{\theta^{t}}$:
\begin{equation*}
    y_{1} \sim \pi_{\theta^{t}}(V, x), ~ y_{2} \sim \pi_{\theta^{t}}(V, x).
\end{equation*}

We then select a better response between two responses by leveraging the current VLMM to evaluate its own responses, \ie, VLMM-as-a-judge.
In particular, we provide the VLMM with the visual context $c_{t}$ for enhanced visual clarity (more detailed in Sec.~\ref{sec:approach:self_ref}).
We can present this preference selection procedure as follows: 
\begin{equation*}
    (y_{w}, y_{l}) \sim  \pi_{\theta^{t}}(V, x, c_{t}, y_{1}, y_{2}),
\end{equation*}
where $y_1$ and $y_2$ are two sampled responses, $y_{w}$ is the chosen response, and $y_{l}$ is the rejected response.

After constructing the preference data at $t$-th iteration as $D_{t}^{\text{pref}} = \{V, x, y_{w}, y_{l}\}$, we use this dataset to perform preference optimization on the current VLMM $\pi_{\theta^{t}}$ using DPO.
The DPO objective for the current VLMM $\pi_{\theta^{t}}$ is represented as follows:
\begin{equation*}
\resizebox{\linewidth}{!}{
    $\begin{aligned}
        &\hspace{1.0cm}\mathcal{L}_{\text{DPO}}(\pi_{\theta^t}; \pi_{\text{ref}, t}) = \\
        &-\mathbb{E}_{(V,x,y_w,y_l) \sim \mathcal{D}_{t-1}^{\text{pref}}} \left[ \log \sigma \left( \beta \log \frac{\pi_{\theta^t}(y_w \mid V,x)}{\pi_{\text{ref},t}(y_w \mid V,x)} \right. \right.\\
        &\hspace{3.7cm} \left. \left. - \beta \log \frac{\pi_{\theta^t}(y_l \mid V,x)}{\pi_{\text{ref}, t}(y_l \mid V,x)} \right) \right],
    \end{aligned}$
}
\end{equation*}
where $\pi_{ref,{t}}$ is the current base reference model, $\beta$ is a hyper-parameter controlling the deviation from the current base reference model and $\sigma$ is the sigmoid function.

\paragraph{Iterative training.}
Our overall iterative training procedure follows previous work~\cite{yuan2024selfrewarding}, where a series of models $\pi_{\theta^1}, \ldots, \pi_{\theta^{T}}$ is trained sequentially.
Each successive model at iteration of $t+1$ uses preference data $D_{t}^{\text{pref}}$ generated by the VLMM at iteration $t$, defined as follows:
\begin{equation*}
        \pi_{\theta^{t+1}}: \text{Training with } D_{t}^{\text{pref}} \text{ initialized from } \pi_{\theta^{t}},
\end{equation*}
where the $t$-th model $\pi_{\theta^{t}}$ generates preference data $D_{t}^{\text{pref}}$ through self-judgment.

\begin{table*}[t]
\centering
\resizebox{0.8\linewidth}{!}{
    \begin{tabular}{lcccccc}
    \toprule
    \multirow{2}{*}{\textbf{Methods}}
    & \multicolumn{2}{c}{\textbf{ActivityNet-QA}} 
    & \multicolumn{2}{c}{\textbf{VIDAL-QA}} 
    & \multicolumn{2}{c}{\textbf{WebVid-QA}} \\
    \cmidrule(lr){2-3}                  
    \cmidrule(lr){4-5}
    \cmidrule(lr){6-7}
    & Acc. & Score 
    & Acc. & Score 
    & Acc. & Score \\
    \midrule
    Video-ChatGPT~\cite{Maaz2023VideoChatGPT}
            & 34.17 & 2.19 & 29.35 & 2.10 & 38.88 & 2.27 \\
    LLaMA-VID~\cite{li2023llamavid}
            & 36.54 & 2.27 & 30.58 & 2.15 & 36.99 & 2.24 \\
    Chat-UniVi~\cite{jin2023chatunivi}
            & 39.35 & 2.32 & 31.40 & 2.16 & 40.05 & 2.31 \\
    Video-LLaVA~\cite{videollava}
            & 41.35 & 2.38 & 34.30 & 2.24 & 42.47 & 2.39 \\
    VLM-RLAIF${^\dagger}$~\cite{vlm_rlaif}
            & 53.27 & 2.56 & 44.82 & 2.40 & 53.69 & 2.62 \\
    PLLaVA${^\dagger}$~\cite{xu2024pllava}
            & 48.44 & 2.50 & 42.45 & 2.39 & 53.55 & 2.59 \\
    LLaVA-NeXT-DPO${^\dagger}$~\cite{zhang2024llavanextvideo}
            & 68.05 & 2.88 & 61.52 & 2.72 & 73.35 & 3.00 \\
    LLaVA-Hound-DPO~\cite{llava-hound}
            & \underline{76.62} & \underline{3.18} & \underline{70.06} & \underline{3.04} & \underline{79.82} & \underline{3.29} \\
    \midrule
    \method ($\pi_{\theta^{1}}$)
            & 75.58 & 3.14 
            & 70.07 & 3.02 
            & 80.74 & 3.28 \\
    \method ($\pi_{\theta^{5}}$)
            & 81.62 & 3.25 
            & 77.33 & 3.10 
            & 86.92 & 3.39 \\
    \method ($\pi_{\theta^{9}}$)
            & \textbf{82.99} & \textbf{3.26} 
            & \textbf{79.00} & \textbf{3.13} 
            & \textbf{88.11} & \textbf{3.40} \\
    \bottomrule
    \end{tabular}
}
\caption{\textbf{Quantitative comparison between different VLMMs on \emph{in-domain} video question answering with detailed captions as supporting evidence proposed in~\citet{llava-hound}.} Our final iterated model of \method ($\pi_{\theta^{9}}$) consistently outperforms all other models in both accuracy and score across these benchmarks, demonstrating superior performance in in-domain video question answering tasks. The best results are \textbf{bold} and the second-best results are \underline{underlined}. ${\dagger}$: reproduced by the authors' implementation. All results except ${\dagger}$ are directly sourced from~\citet{llava-hound}.}
% \vspace{-0.5em}
\label{tab:main_quan_in_domain}
\end{table*}

\begin{table*}[!t]
\centering
\resizebox{0.9\linewidth}{!}{
    \begin{tabular}{lcccccccc}
    \toprule
    \multirow{2}{*}{\textbf{Methods}}
    & \multicolumn{2}{c}{\textbf{MSVD-QA}} 
    & \multicolumn{2}{c}{\textbf{MSRVTT-QA}} 
    & \multicolumn{2}{c}{\textbf{TGIF-QA}} 
    & \multicolumn{2}{c}{\textbf{SSV2-QA}} \\
    \cmidrule(lr){2-3}                  
    \cmidrule(lr){4-5}
    \cmidrule(lr){6-7}
    \cmidrule(lr){8-9}
    & Acc. & Score 
    & Acc. & Score 
    & Acc. & Score 
    & Acc. & Score \\
    \midrule
    Video-ChatGPT~\cite{Maaz2023VideoChatGPT}
            & 34.06 & 2.20 & 25.65 & 1.98 & 31.35 & 2.09 & 19.36 & 1.75 \\
    LLaMA-VID~\cite{li2023llamavid}
            & 34.14 & 2.21 & 25.02 & 1.99 & 27.18 & 2.00 & 22.16 & 1.84 \\
    Chat-UniVi~\cite{jin2023chatunivi}
            & 35.61 & 2.23 & 25.89 & 2.01 & 33.23 & 2.13 & 20.59 & 1.79 \\
    Video-LLaVA~\cite{videollava}
            & 39.46 & 2.37 & 30.78 & 2.15 & 32.95 & 2.18 & 24.31 & 1.90 \\
    VLM-RLAIF${^\dagger}$~\cite{vlm_rlaif}
            & 51.16 & 2.55 & 41.44 & 2.30 & 46.52 & 2.41 & 29.78 & 1.94 \\
    PLLaVA${^\dagger}$~\cite{xu2024pllava}
            & 48.92 & 2.53 & 38.26 & 2.28 & 43.83 & 2.40 & 30.92 & 2.07 \\
    LLaVA-NeXT-DPO${^\dagger}$~\cite{zhang2024llavanextvideo}
            & 65.08 & 2.82 & 59.12 & 2.65 & 60.80 & \underline{2.70} & 40.14 & 2.24 \\
    LLaVA-Hound-DPO~\cite{llava-hound}
            & \underline{73.64} & \underline{3.12} & \underline{68.29} & \underline{2.98} & \underline{74.00} & \textbf{3.12} & \underline{48.89} & \underline{2.53} \\
    \midrule
    \method ($\pi_{\theta^{1}}$)
            & 74.33 & 3.12 & 68.18  & 2.96 & 73.57 & 3.10 & 48.91 & 2.52 \\
    \method ($\pi_{\theta^{5}}$)
            & 79.63 & 3.19 & 74.07  & 3.05 & 77.52 & 3.12 & 53.13 & 2.57 \\
    \method ($\pi_{\theta^{9}}$)
            & \textbf{80.36} & \textbf{3.20} & \textbf{75.42}  & \textbf{3.05} & \textbf{78.58} & \textbf{3.12} & \textbf{54.66} & \textbf{2.59} \\
    \bottomrule
    \end{tabular}
}
\caption{\textbf{Quantitative comparison between different VLMMs on \emph{out-domain} video question answering with detailed captions as supporting evidence proposed in~\citet{llava-hound}.} Our final iterated model of \method ($\pi_{\theta^{9}}$) consistently outperforms all other models in both accuracy and score across these benchmarks, demonstrating superior performance in out-domain video question answering tasks. The best results are \textbf{bold} and the second-best results are \underline{underlined}. ${\dagger}$: reproduced by the authors' implementation. All results except ${\dagger}$ are directly sourced from~\citet{llava-hound}.}
% \vspace{-1em}
\label{tab:main_quan_out_domain}
\end{table*}

\subsection{Self-Retrospective Preference Modeling}
\label{sec:approach:self_ref}
A key aspect of iterative DPO in VLMM involves using a VLMM as a judge to iteratively select preferences that accurately answer posed questions~\cite{vlm_rlaif}.
Specifically, we provide the VLMM with detailed visual descriptions as visual context, generated by the VLMM itself in addition to the video content for improved visual clarity.
Moreover, inspired by humans learning process, we enhance the visual context in a \emph{self-retrospective} manner.
Just as retrospection allows humans to make better decisions by reflecting on the past~\cite{simon1962architecture,madaan2023selfrefine}, we leverage previously generated visual context to generate better context, enhancing the accuracy and relevance of the preference selection process, defined as follows:
\begin{equation*}
    c_{t} \sim \pi_{\theta^{t}}(V, c_{t-1}),
\end{equation*}
where $c_{t-1}$ is the previous visual context at time $t-1$.

Using the generated context $c_{t}$, question $x$, video $V$, and responses $\{y_{1}, y_{2}\}$, we classify the chosen $y_{w}$ and rejected data $y_{l}$ from responses using the current aligned VLMM $\pi_{\theta^{t}}$, a process we call \emph{self-retrospective} preference modeling, thereby constructing preference data $D_{t}^{\text{pref}}$ at time $t$.

\begin{table*}[t]
\centering
\resizebox{0.75\linewidth}{!}{
    \begin{tabular}{lcccccc}
    \toprule
    \multirow{2}{*}{\textbf{Methods}}
    & \multicolumn{2}{c}{\textbf{MSVD-QA}}
    & \multicolumn{2}{c}{\textbf{MSRVTT-QA}}
    & \multicolumn{2}{c}{\textbf{TGIF-QA}} \\
    \cmidrule(lr){2-3}
    \cmidrule(lr){4-5}
    \cmidrule(lr){6-7}
    & Acc. & Score 
    & Acc. & Score 
    & Acc. & Score \\
    \midrule
    Video-ChatGPT~\cite{Maaz2023VideoChatGPT}
            & 68.6 & 3.8 & 58.9 & 3.4 & 47.8 & 3.2 \\
    Chat-UniVi~\cite{jin2023chatunivi}
            & 70.0 & 3.9 & 53.1 & 3.1 & 46.1 & 3.1 \\
    VideoChat2~\cite{videochat2}
            & 70.0 & 3.9 & 54.1 & 3.3 & - & - \\
    Video-LLaVA~\cite{videollava}
            & 71.8 & 3.9 & 59.0 & 3.4 & 48.4 & 3.2 \\
    LLaMA-VID~\cite{li2023llamavid}
            & 72.6 & 3.9 & 58.7 & 3.4 & 49.2 & 3.3 \\
    PLLaVA${^\dagger}$~\cite{xu2024pllava}
            & 78.8 & 4.0 & 65.6 & 3.4 & 57.9 & \textbf{3.5} \\
    LLaVA-NeXT-DPO${^\dagger}$~\cite{zhang2024llavanextvideo}
            & 78.6 & 4.0 & 63.4 & 3.1 & 58.2 & \underline{3.4} \\
    VLM-RLAIF${^\dagger}$~\cite{vlm_rlaif}
            & \underline{81.0} & \underline{4.2} & 69.2 & \underline{3.7} & \underline{62.3} & \textbf{3.5} \\
    LLaVA-Hound-DPO~\cite{llava-hound}
            & 80.7 & 4.1 & \underline{70.2} & \underline{3.7} & 61.4 & \textbf{3.5} \\
    \midrule
    \method ($\pi_{\theta^{1}}$)
            & 80.1 & 4.1 
            & 69.8 & 3.6 
            & 61.0 & 3.4 \\
    \method ($\pi_{\theta^{5}}$)
            & 84.8 & 4.3 
            & 76.0 & 3.8 
            & 66.8 & 3.5 \\
    \method ($\pi_{\theta^{9}}$)
            & \textbf{85.8} & \textbf{4.3} 
            & \textbf{78.7} & \textbf{3.9} 
            & \textbf{67.8} & \textbf{3.5} \\
    \bottomrule
    \end{tabular}
}
\caption{\textbf{Comparison of different VLMMs on \emph{out-domain} video question answering benchmark~\citep{Maaz2023VideoChatGPT}.} \method ($\pi_{\theta^{9}}$) outperforms previous work across three video question answering datasets. 
Best results in \textbf{bold}, second-best \underline{underlined}.
${\dagger}$: reproduced with the authors' implementation.
eOther results are directly sourced from~\citet{llava-hound}.}
\label{tab:main_quan_brief}
\end{table*}

%%%%%%%%%%%%%%%%%%%%%%%%%%%%%%%%%%%%%%%%%%%%%%%%%%%%%%%%%%%%%%%%%%%%%%%%%%%%%%%%%%%
\section{Experiments}
\label{sec:exp}

\subsection{Experimental Setup}
\label{sec:exp_su}

\paragraph{Dataset details.}
Our training dataset utilizes a fixed set of 17k video-instruction  ($\{V, x\}$) pairs from~\cite{llava-hound}, in contrast to previous works~\cite{yuan2024selfrewarding,chen2024selfplay} that incremented their dataset across iterations.
For all iterations beyond the initial VLMM $\pi_{\theta^{1}}$, we generate preference dataset $D_{t}^{\text{pref}}$ at each iteration by generating new responses and preferences. 
Following ~\cite{Maaz2023VideoChatGPT,llava-hound}, we evaluate our method on two types of video question answering datasets: one that requires concise responses, and the other that demands comprehensive answers, across 7 video collections.

\paragraph{Training details.}
We perform full-parameter fine-tuning using DPO with 9 total iterations, tripling the previous iterative preference optimization approach for LLMs alignment~\cite{yuan2024selfrewarding}. 
All generative processes use specific prompts.
Training is conducted on 8$\times$NVIDIA A100 GPUs (80G). 
We employ a 7B-sized model for fair comparison with others.

\subsection{Quantitative Analysis}
\label{sec:exp:quan} 

\paragraph{In-domain video question answering.}
As shown in Tab.~\ref{tab:main_quan_in_domain}, \method demonstrates consistent performance gains at each iteration up to the 9$th$ iteration. 
Moreover, final iterated model ($\pi_{\theta^{9}}$), outperforms all previous work across all video benchmarks in both accuracy and score by a noticeable margin.
We attribute this performance improvement to the better alignment of video modality provided by the proposed iterative retrospective judgment for VLMMs.

\paragraph{Out-domain video question answering.}
For evaluating out-domain video question answering, we use two types of datasets. 
Tables~\ref{tab:main_quan_out_domain} and~\ref{tab:main_quan_brief} show the comparative results for datasets that require complex answers and concise keyword answers, respectively. 
The final iterated model of~\method ($\pi_{\theta^{9}}$) outperforms the previous work by a large margin in both cases, demonstrating its effectiveness in generating both detailed and precise responses. 
This model also shows consistent performance improvements at each iteration, as shown in Tables~\ref{tab:main_quan_out_domain} and~\ref{tab:main_quan_brief}.

\begin{figure}[!t]
    \centering
    \includegraphics[width=1.0\linewidth]{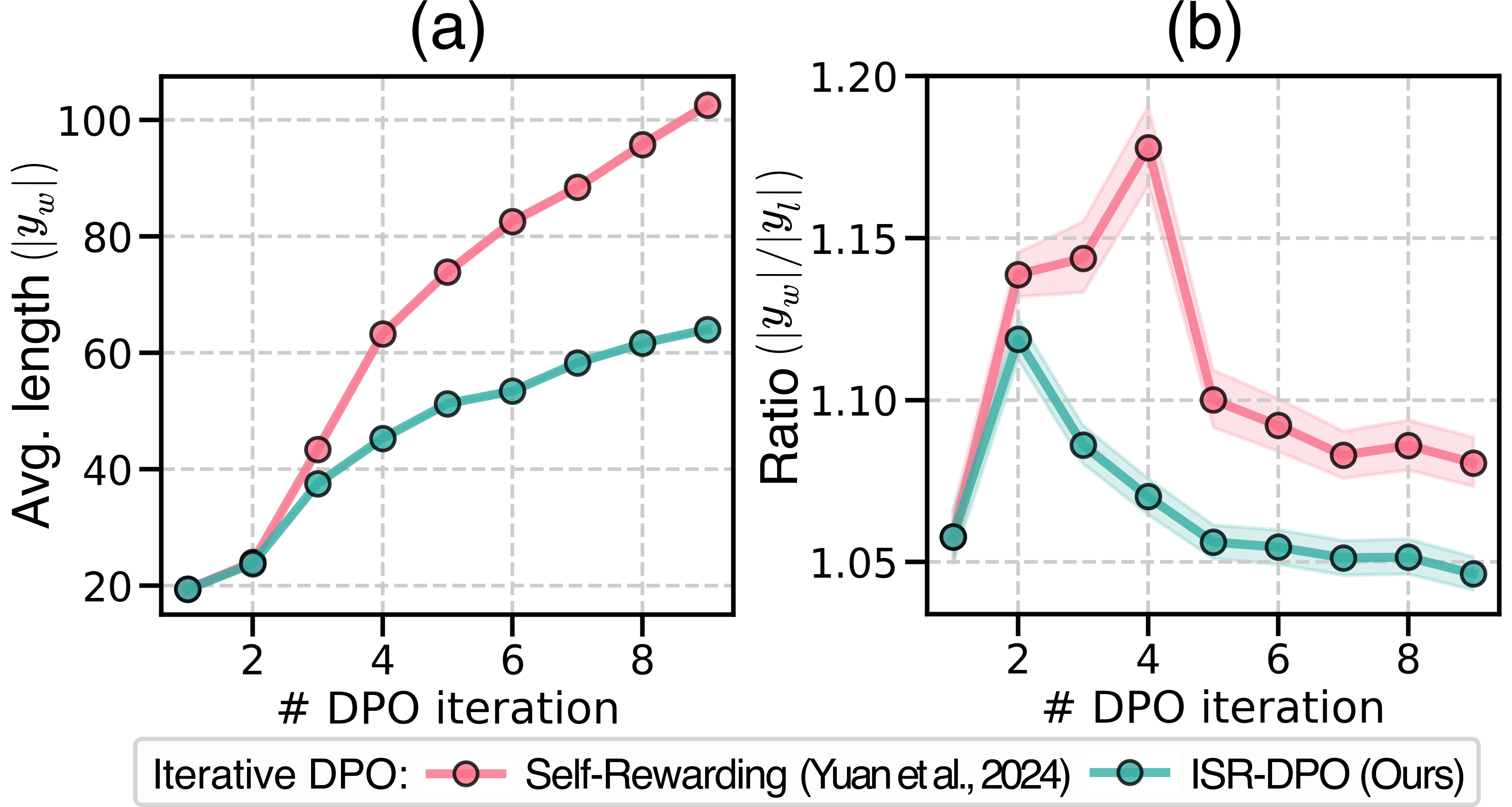}
    \caption{\textbf{Length analysis of preference dataset during iterative DPO.} (a) Average (Avg.) word length of chosen response $|y_{w}|$ in preference dataset $D_{t}^{\text{pref}}$ across DPO iterations.
    Self-rewarding results in longer responses compared to the \method. (b) Ratio of the word lengths of chosen responses ($|y_{w}|$) to rejected responses ($|y_{l}|$). \method consistently maintains a lowered ratio compared to the self-rewarding, indicating reduced response length after optimized.
    `\# DPO iteration' means the number of DPO iterations.
    }
    \label{fig:length_analysis_in_pref}
\end{figure}

\begin{figure*}[t]
    \centering
    \includegraphics[width=0.95\linewidth]{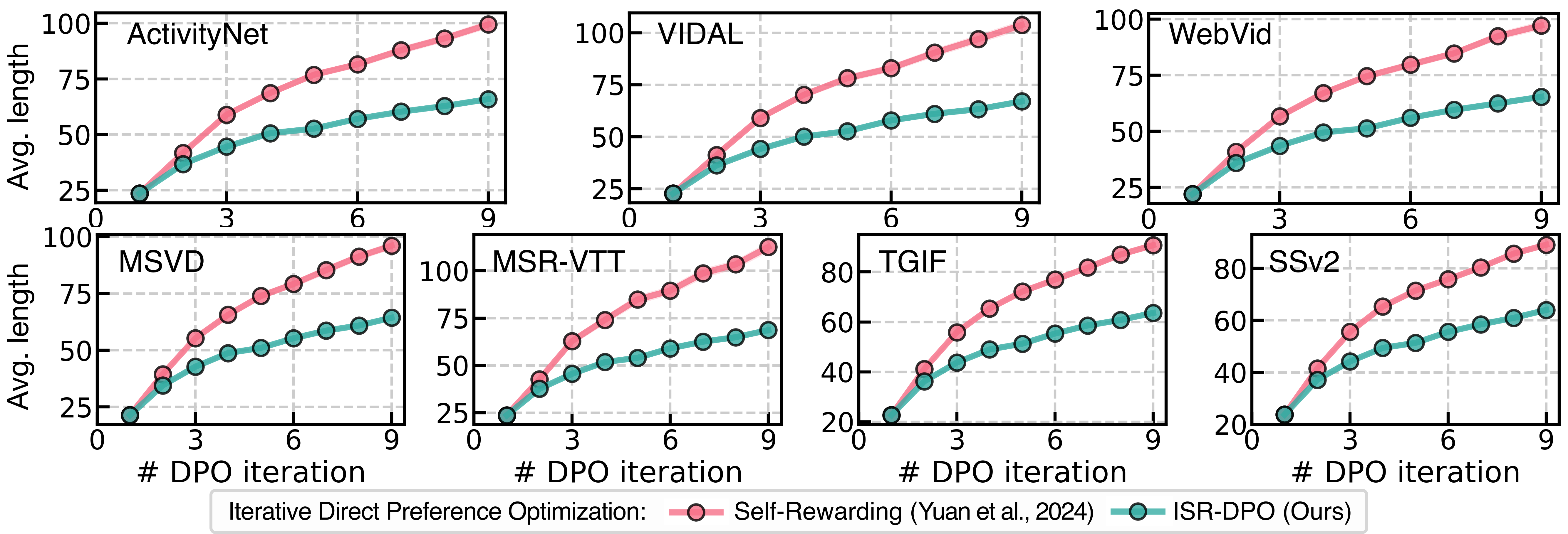}
    % \vspace{-1.5em}
    \caption{\textbf{Average (Avg.) response word length between self-rewarding and \method on various video question answering benchmarks.} \method yields compact and concise responses at the same iteration compared to self-rewarding.
    }
    \label{fig:fig_len_comp}
    \vspace{-0.5em}
\end{figure*}

\begin{figure}[!t]
    \centering
    \includegraphics[width=0.96\linewidth]{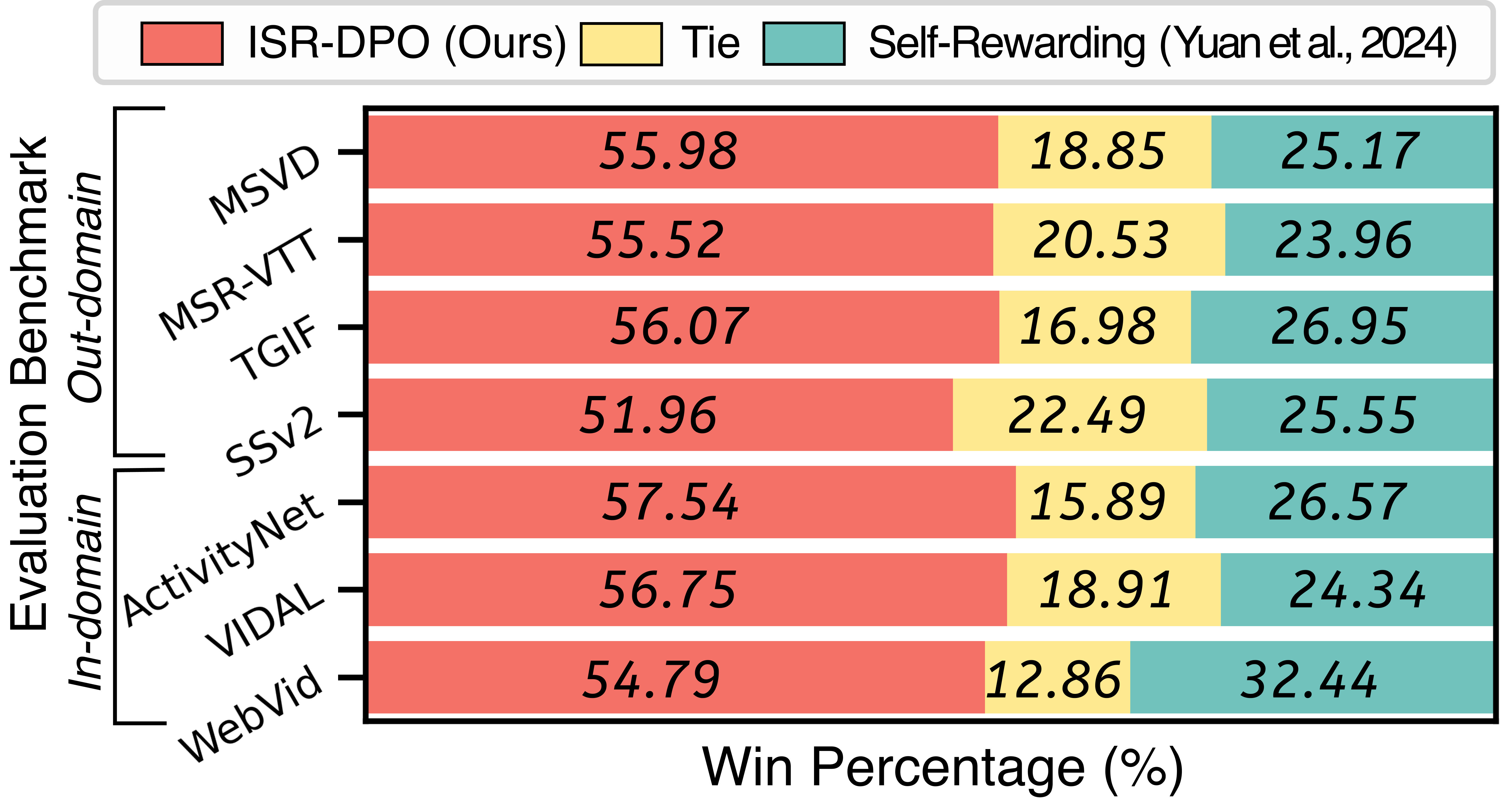}
    \caption{\textbf{Head-to-head performance comparison at 9th iteration.} 
    \method consistently outperforms the self-rewarding across benchmarks. 
    }
    \label{fig:win_ratio}
    \vspace{-0.5em}
\end{figure}

\subsection{Detailed Analysis}
\label{sec:exp:da}
To evaluate the effectiveness of \method, we address the following research questions, specifically exploring the effect and design of visual context:
% \vspace{-0.5em}
 \begin{itemize}
\setlength\itemsep{-0.3em}
    \item \textbf{RQ1}: What are the effects and benefits of visual context during iterative DPO?
    \item \textbf{RQ2}: How should the visual context be designed?
    % \vspace{-1.0em}
\end{itemize}
In particular, we compare \method with self-rewarding~\cite{yuan2024selfrewarding}, which serves as our baseline for adopting iterative DPO in VLMMs without self-retrospective context.

\begin{table}[!t]
\centering
\resizebox{1.0\linewidth}{!}{
    \begin{tabular}{lcc}
    \toprule 
    \multirow{2}{*}{Task}
    & \multicolumn{2}{c}{\textbf{Human Alignment Accuracy (\%)}} \\
    \cmidrule{2-3}
    & Self-rewarding & \method \\
    \midrule
    Preference selection
            & 59.0 & 75.0 \\
    \bottomrule
    \end{tabular}
}
\caption{\textbf{Human annotator alignment accuracy for preference selection.} We measure human alignment accuracy to evaluate the amount of correlation between human and aligned models, \ie, self-rewarding \vs \method.}
\label{tab:pref_human_eval}
\end{table}

\begin{table*}[t]
\centering
\resizebox{1.0\linewidth}{!}{
    \begin{tabular}{lcccccccccccccc}
    \toprule
    \multirow{3}{*}{\specialcell{\textbf{Context}\\\textbf{Design}}}
    & \multicolumn{8}{c}{\textbf{Out-of-domain Video QA Benchmark}}
    & \multicolumn{6}{c}{\textbf{In-domain Video QA Benchmark}} \\
    \cmidrule(lr){2-9}
    \cmidrule(lr){10-15}
    & \multicolumn{2}{c}{\textbf{MSVD}} 
    & \multicolumn{2}{c}{\textbf{MSRVTT}} 
    & \multicolumn{2}{c}{\textbf{TGIF}} 
    & \multicolumn{2}{c}{\textbf{SSV2}} 
    & \multicolumn{2}{c}{\textbf{ActivityNet}} 
    & \multicolumn{2}{c}{\textbf{VIDAL}} 
    & \multicolumn{2}{c}{\textbf{WebVid}} \\
    \cmidrule(lr){2-3}                  
    \cmidrule(lr){4-5}
    \cmidrule(lr){6-7}
    \cmidrule(lr){8-9}
    \cmidrule(lr){10-11}
    \cmidrule(lr){12-13}
    \cmidrule(lr){14-15}
    & Acc. & Score 
    & Acc. & Score 
    & Acc. & Score 
    & Acc. & Score 
    & Acc. & Score 
    & Acc. & Score 
    & Acc. & Score \\
    \midrule
    \specialcell{N/A}
            & 78.73 & 3.14 & 73.42 & 3.00 & 77.10 & 3.09 & 54.34 & 2.56 
            & 81.96 & 3.23 & 76.71 & 3.09 & 87.24 & 3.39 \\
    Fixed
            & 79.17 & 3.15 & 74.35 & 3.02 & 77.88 & 3.09 & 54.29 & 2.57 
            & 82.25 & 3.24 & 77.90 & 3.12 & 87.49 & 3.39 \\
    Renew
            & 79.49 & 3.19 & 74.04 & 3.04 & 77.63 & 3.12 & 53.03 & 2.56 
            & 82.03 & 3.26 & 77.73 & 3.12 & 86.68 & 3.38 \\
    \specialcell{Retrospective}
            & \textbf{80.36} & \textbf{3.20} & \textbf{75.42} & \textbf{3.05} & \textbf{78.58} & \textbf{3.12} & \textbf{54.66} & \textbf{2.59} 
            & \textbf{82.99} & \textbf{3.26} & \textbf{79.00} & \textbf{3.13} & \textbf{88.11} & \textbf{3.40} \\
    \bottomrule
    \end{tabular}
}
\caption{\textbf{Quantitative comparison of various designs for generating visual context.} `N/A' indicates no use of context, `Fixed' uses context generated in the first iteration for all subsequent iterations, `Renew' generates new context each iteration, and `Retrospective.' employs a self-retrospective context.}
\label{tab:main_quan_context_method}
\end{table*}

\subsubsection{Effect of visual context during iterative process}
\label{sec:da:visual_context}
Figure~\ref{fig:length_analysis_in_pref} demonstrates the effect of including visual context during preference selection.
As shown in Fig.~\ref{fig:length_analysis_in_pref}-(a), \method generates shorter chosen responses compared to self-rewarding as training iterations progress. 
Similarly, Fig.~\ref{fig:length_analysis_in_pref}-(b) shows a lower ratio of chosen to rejected response lengths in \method.
We posit that dual conditioning on video content and visual context during preference selection enables the VLMM to select preferences based on video information rather than length bias.
This results in a lower chosen-to-rejected preference ratio and shorter, more concise responses from the VLMM, as illustrated in Fig.~\ref{fig:fig_len_comp}.

Moreover, we compare the 9$th$ iteration model's responses between self-rewarding and \method to validate the effectiveness of visual context, as in \citet{yuan2024selfrewarding}.
In particular, we use GPT-4 as the evaluator by selecting the response closest to the ground truth, assessing win-rates.
% In particular, we use GPT-4 as the evaluator by selecting the response closest to the ground truth, assessing win-rates (details in supplementary).
Figure~\ref{fig:win_ratio} shows the win-rate between self-rewarding and \method across all benchmarks, demonstrating the effectiveness of \method.
Notably, despite generating more concise responses (Fig.~\ref{fig:fig_len_comp}), \method consistently achieved higher winning rates across all benchmarks.
This provides evidence of \method's effectiveness at conveying more relevant and accurate information within concise responses, mitigating verbosity hallucinations.

\paragraph{Effect of visual context on human alignment.}
To evaluate the impact of visual context on judgment quality, we assess the correspondence between AI models' preferences and those of human annotators, following~\citet{lee2023rlaif}.
As shown in Tab.~\ref{tab:pref_human_eval}, \method demonstrates a higher human alignment accuracy (75.0 \%) compared to self-rewarding (59.0 \%), suggesting that the incorporation of visual context enhances the model's ability to make human-like assessments.

\subsubsection{Various design choices for visual context.}
\label{sec:da:self_retro}
\begin{figure}[!t]
    \centering
    \includegraphics[width=\linewidth]{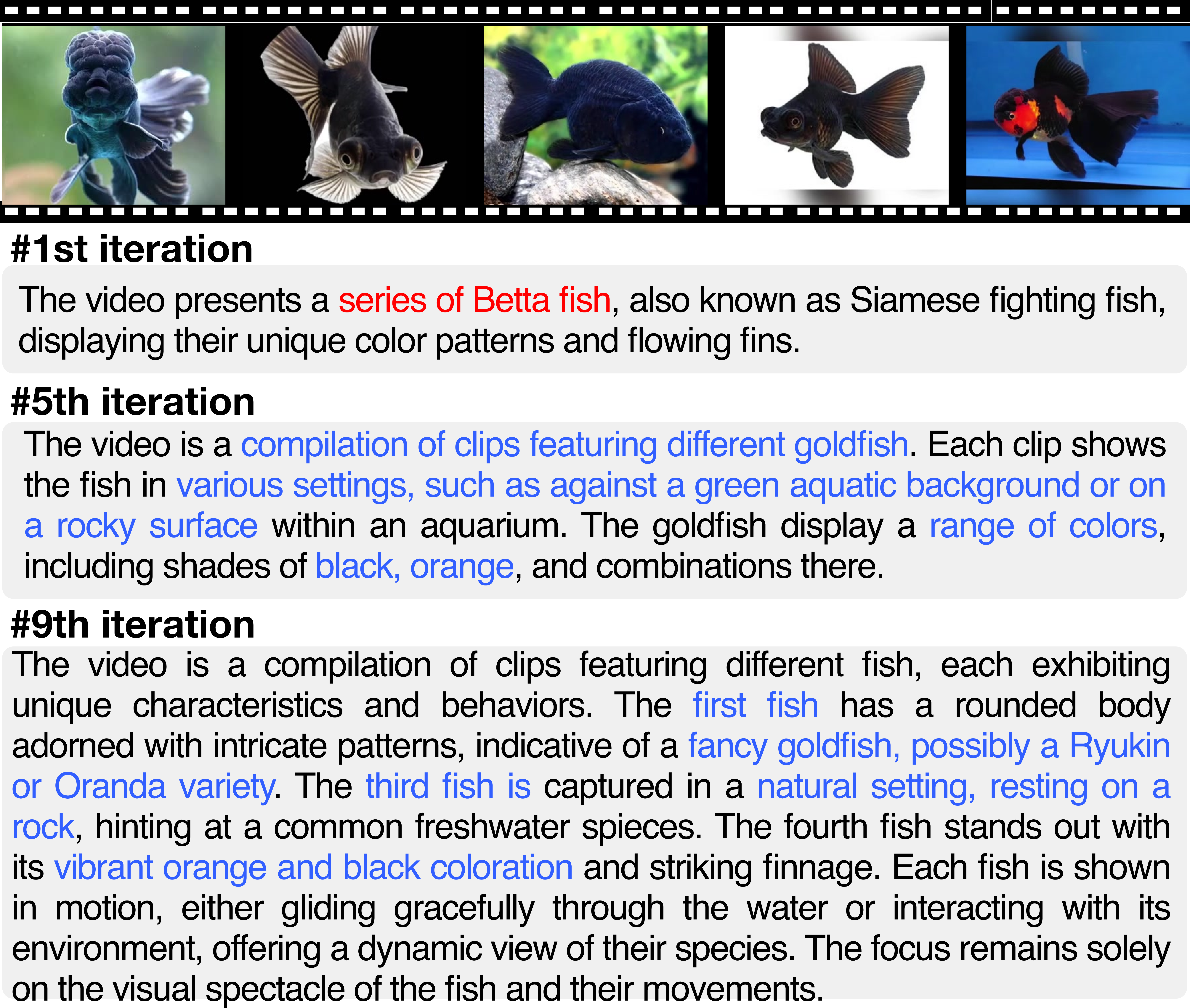}
    \caption{\textbf{Visualization of predicted context over iteration.} 
    Generated context becomes increasingly well-grounded over iteration.
    \textcolor{red}{Red} indicates irrelevant responses, while \textcolor{blue}{blue} indicates accurate, visually-grounded responses.
    }
    \label{fig:context_comp}
    \vspace{-0.5em}
\end{figure}

We examine various design choices for visual context in Tab.~\ref{tab:main_quan_context_method}: (1) without context (`N/A'), (2) Fixed context from the first iteration (`Fixed'), (3) New context at each iteration (`Renew') and (4) Self-retrospective context (`Self-retro.'). 
The `Self-retro.' consistently performs the best, leveraging and refining previous context while adding details with improved video understanding (Fig.~\ref{fig:context_comp}).
Interestingly, `Fixed' outperforms `Renew' in most benchmarks, except for MSRVTT.
For SSv2 and WebVid, `Renew' even performs worse than `N/A'.
We hypothesize that `Renew' may introduce inconsistent focus in the video across iterations, potentially causing attention to irrelevant details.
These findings suggest that a methodical approach to context renewal, such as our `Self-retro.', is crucial for maintaining focus on relevant content, thereby improving proper preference modeling.

\begin{figure}[!t]
    \centering
    \includegraphics[width=1\linewidth]{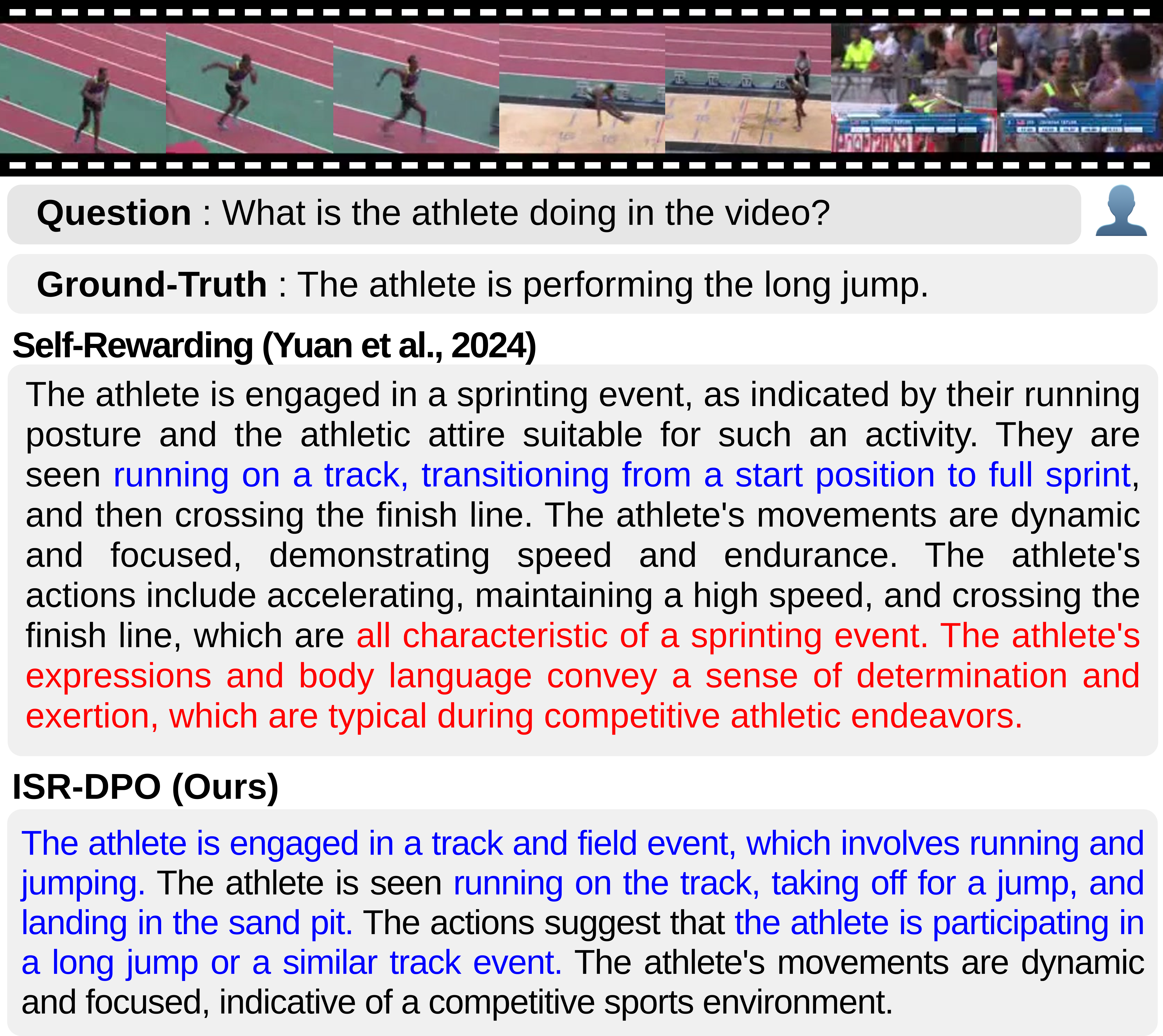}
    \caption{\textbf{Qualitative comparison of self-rewarding \vs \method.} The figure contrasts descriptions generated without visual context, \ie, self-rewarding (upper), against those with visual context, \ie, \method (bottom), at the 9$th$ iteration. Visual context results in more accurate, concise, and relevant descriptions. \textcolor{red}{Red} indicates irrelevant responses, while \textcolor{blue}{blue} indicates well-grounded responses.
    }
    \label{fig:perf_on_iter}
    \vspace{-0.5em}
\end{figure}

\subsection{Qualitative Analysis}
\label{sec:quali}
\paragraph{Enhanced visual context over iteration.}
To show the improving nature of self-retrospective context, we visualize the generated context as shown in Fig.~\ref{fig:context_comp}.
As training iterations progress, the context adds more and more detailed visual information about the video, such as specific species of goldfish.
This improved context aids the overall understanding of the video content to improve preference selection process.

\paragraph{Comparison of self-rewarding \vs \method.}
Figure~\ref{fig:perf_on_iter} compares the responses of self-rewarding, \ie, \method w/o visual context, and \method for 9$th$ iterated models.
The self-rewarding tends to produce longer responses, but as sentences progress, they become less relevant to the question and visual content. 
Also, it fails to recognize the athlete's jumping motion accurately.
In contrast, \method generates more concise and accurate responses that are well-grounded in the video content.

%%%%%%%%%%%%%%%%%%%%%%%%%%%%%%%%%%%%%%%%%%%%%%%%%%%%%%%%%%%%%%%%%%%%%%%%%%%%%%%%%%%
\section{Conclusion}
We present \method, a novel iterative direct preference optimization for VLMMs that enhances the instruction-following ability for videos.
In particular, we propose self-retrospective preference modeling to improve VLMM's capability to judge visually grounded preferences.
By doing so, \method mitigates the model's problematic inclination for visually ungrounded verbosity in judging preferred response, leading to more concise and visually grounded responses.
Empirical evaluations across various video question answering benchmarks demonstrate \method's superior performance compared to the state of the art VLMMs.

\section{Acknowledgment}
This work was partly supported by CARAI grant funded by DAPA and ADD (UD230017TD) and the IITP grants (No.RS-2022-II220077, No.RS-2022-II220113, No.RS-2022-II220959, No.RS-2022-II220871, No.RS-2021-II211343 (SNU AI), No.RS-2021-II212068 (AI Innov. Hub)) funded by the Korea government(MSIT).

\bibliography{aaai25}

% \appendix
\section{Additional Input Prompts for Preference Dataset Generation}
\label{sec:appendix:prompt_pref}
In the process of generating our preference dataset, we employ specific additional input prompts for each stage.
Figure \ref{fig:appendix_prompts} illustrates three types of input prompts used in this process: 1) response generation, 2) self-retrospective context generation, and 3) preference judgment.
The `Prompt (response)' defines the guideline for VLMM's responses and is used consistently throughout all stages of data generation.
The `Prompt (context)' demonstrates the prompt used to generate a context based on the previous context.
Lastly, the `Prompt (judge)' presents the prompt used for preference judgment using the current Video Large Multimodal Model (VLMM).

\section{Details on Head-to-Head Comparison with GPT-4 Evaluator}
\label{sec:appendix:gpt4_eval}
We evaluate the generated response quality of the~\method compared to self-rewarding~\cite{yuan2024selfrewarding} through a head-to-head comparison~\cite{yuan2024selfrewarding}.
Specifically, we prompted GPT-4 to determine which of the two responses is superior across in-domain and out-of-domain video question answering benchmarks.
The evaluation focused on two key aspects: 1) the relevance of model's answer to the provided instruction, and 2) the accuracy of the response in relation to the ground-truth answer.
We visualize a detailed prompt in Fig.~\ref{fig:prompt_gpt_win_rate}.

\section{Details on Human Evaluation for Human Preference Alignment}
\label{sec:appendix:human_eval}
% Following~\cite{lee2023rlaif}, w
We conduct a human evaluation to measure how well the AI-generated preferences align with human preferences, following the approach of ~\citet{lee2023rlaif}. 
We randomly sample 100 questions from the validation set of video question-answering dataset~\cite{xu2017video}. 
We then recruit 15 annotators per question through the Amazon Mechanical Turk platform. 
Annotators are presented with a video, an instruction, and two versions of responses generated from our \method.
Specific instructions and examples of the questions given to the annotators can be found in Fig.~\ref{fig:mturk_instruction}.

\section{More Qualitative Results}
\label{sec:appendix:more_qual}
In Fig. \ref{fig:quali_comp_w_wo_ctx}, we present additional examples comparing responses generated by self-rewarding and our \method.
Well-grounded phrases are highlighted in blue, while misaligned or irrelevant phrases are marked in red.
Compared to self-rewarding, our approach reduces the occurrence of misaligned and overly verbose sentences.
For instance, in the beach soccer example, our method accurately identifies the team colors as blue and orange without unnecessary elaboration.
These examples demonstrate how our \method reduces verbosity hallucination, generating more concise and relevant responses.

\section{Performance Over Training Iterations}
\begin{figure*}[t]
    \centering
    \includegraphics[width=\linewidth]{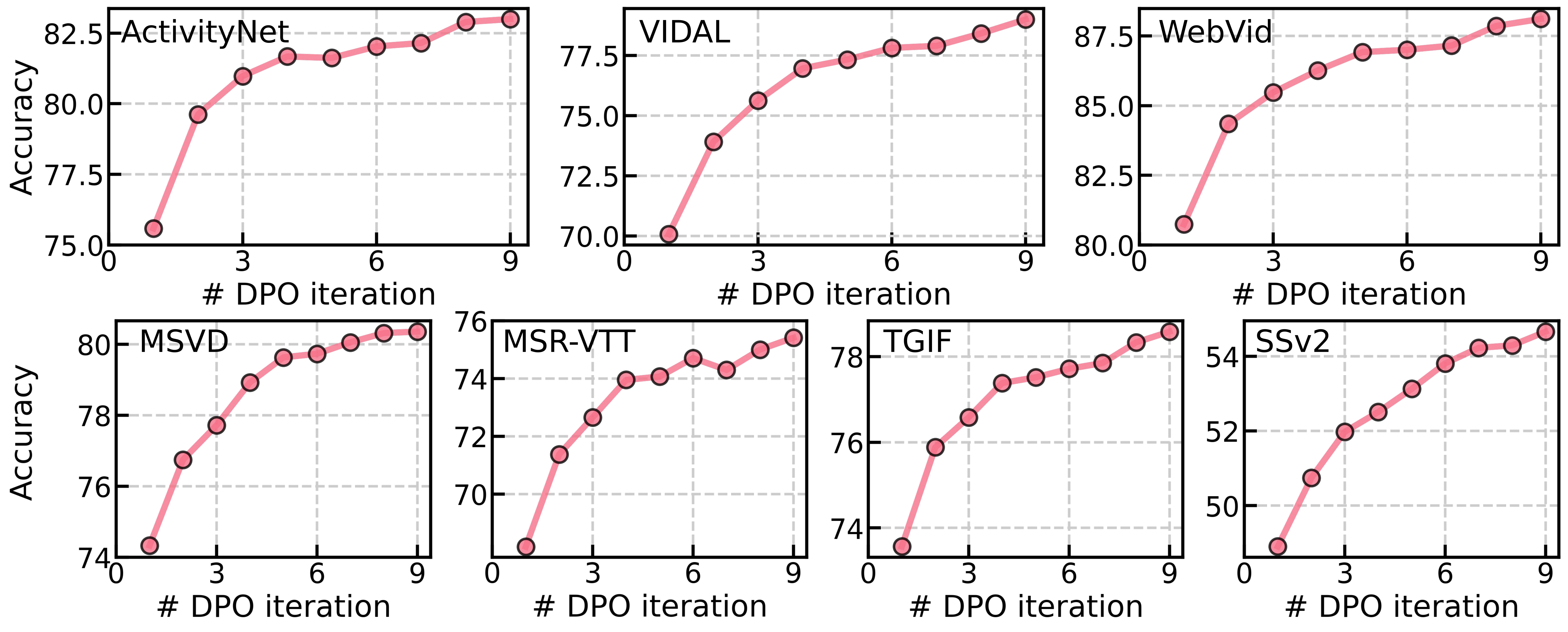}
    \caption{\textbf{Accuracy of \method over iterations on video question answering benchmarks.} Overall, our \method consistently improves its performance over DPO iteration. In-domain datasets: Activity-Net, VIDAL and WebVid, Out-domain datasets: MSVD, MSR-VTT, TGIF and SSv2 used in~\citet{llava-hound}.}
    \label{fig:appendix_acc_iter}
\end{figure*}

In Fig.~\ref{fig:appendix_acc_iter}, we demonstrate the effectiveness of \method across training iterations using various video question answering benchmarks for evaluation.
Overall, the performance improves as we increase the number of training iterations, with the exception of the MSR-VTT dataset at the 7th iteration.
However, we can observe that the performance recovers and improves again in subsequent training iterations up to the 9th iteration.

% ================================================================================================================================================
% ================================================================================================================================================

\begin{figure*}[t]
    \centering
    \includegraphics[width=\linewidth]{figures/supp_prefdatagen_prompts.pdf}
    \caption{\textbf{Various input prompts for constructing preference dataset.} This shows various input prompts: the upper part for generating two responses, the center part for context generation based on previous context, and the bottom part for preference judgment using the VLMM from the latest iteration.
    }
    \label{fig:appendix_prompts}
\end{figure*}

\begin{figure*}[t]
    \centering
    \includegraphics[width=0.85\linewidth]{figures/supp_gpt_win_rate_prompt.pdf}
    \caption{\textbf{Evaluation criteria provided to GPT-4.} To compare the generated responses of self-rewarding and \method, %~\ref{fig:win_ratio},
    we prompted GPT-4 to choose better response regrading two criteria: Relevance and precision.
    }
    \label{fig:prompt_gpt_win_rate}
\end{figure*}

\begin{figure*}[t]
    \centering
    \includegraphics[width=0.75\linewidth]{figures/supp_mturk_instruction.pdf}
    \caption{\textbf{Evaluation criteria provided to Amazon Mechanical Turk annotators.} We carefully instructed the annotators to penalize the outputs that include unaligned contents with the provided video, or the answer that contains overly verbose sentences that deviates from the question's purposes.
    }
    \label{fig:mturk_instruction}
\end{figure*}

\begin{figure*}[!t]
    \centering
    \includegraphics[width=\linewidth]{figures/supp_9th_comp_pred_per_context.pdf}
    \caption{\textbf{More qualitative example of prediction from self-rewarding \vs \method.} We compare responses generated at the 9\textit{th} iteration for both models. Integrating visual context leads to more accurate, concise, and relevant descriptions that align more closely with the ground-truth answer.
    \textcolor{red}{Red} indicates irrelevant or wrong responses, while \textcolor{blue}{blue} indicates well-grounded responses.
    }
    \label{fig:quali_comp_w_wo_ctx}
    \vspace{-1em}
\end{figure*}

\end{document}